\documentclass[letterpaper, 10 pt, conference]{ieeeconf}  

\IEEEoverridecommandlockouts                              

\usepackage{graphicx}
\usepackage{amsmath}
\usepackage{amssymb}
\usepackage{booktabs}
\usepackage{comment}
\usepackage{multirow}

\title{\LARGE \bf
Approximate Supervised Object Distance Estimation on \\ Unmanned Surface Vehicles
}

\author{Benjamin Kiefer$^{12}$, Yitong Quan$^{2}$ and Andreas Zell$^{2}$
\thanks{$^{1}$LOOKOUT, $^{2}$Cognitive Systems Group,
        University of Tuebingen
        {\tt\small prename.surname@uni-tuebingen.de}}%
}

\begin{document}

\maketitle
\thispagestyle{empty}
\pagestyle{empty}

\begin{abstract}

Unmanned surface vehicles (USVs) and boats are increasingly important in maritime operations, yet their deployment is limited due to costly sensors and complexity. LiDAR, radar, and depth cameras are either costly, yield sparse point clouds or are noisy, and require extensive calibration. Here, we introduce a novel approach for approximate distance estimation in USVs using supervised object detection. We collected a dataset comprising images with manually annotated bounding boxes and corresponding distance measurements. Leveraging this data, we propose a specialized branch of an object detection model, not only to detect objects but also to predict their distances from the USV. This method offers a cost-efficient and intuitive alternative to conventional distance measurement techniques, aligning more closely with human estimation capabilities. We demonstrate its application in a marine assistance system that alerts operators to nearby objects such as boats, buoys, or other waterborne hazards.

\end{abstract}

\section{Introduction}
\label{sec:introduction}


\begin{figure}
\centering
\includegraphics[width=0.49\textwidth]{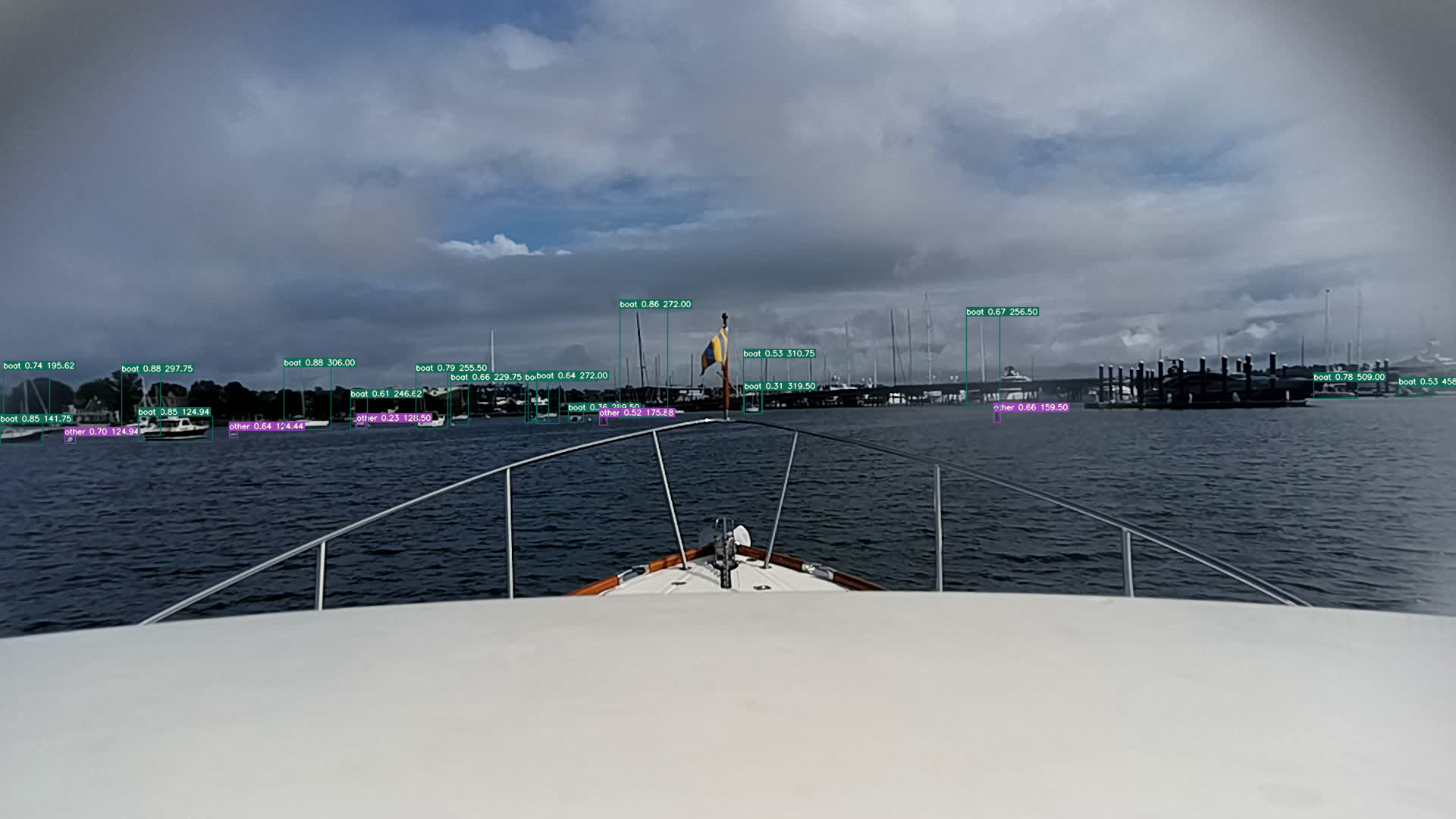}
\includegraphics[width=0.49\textwidth]{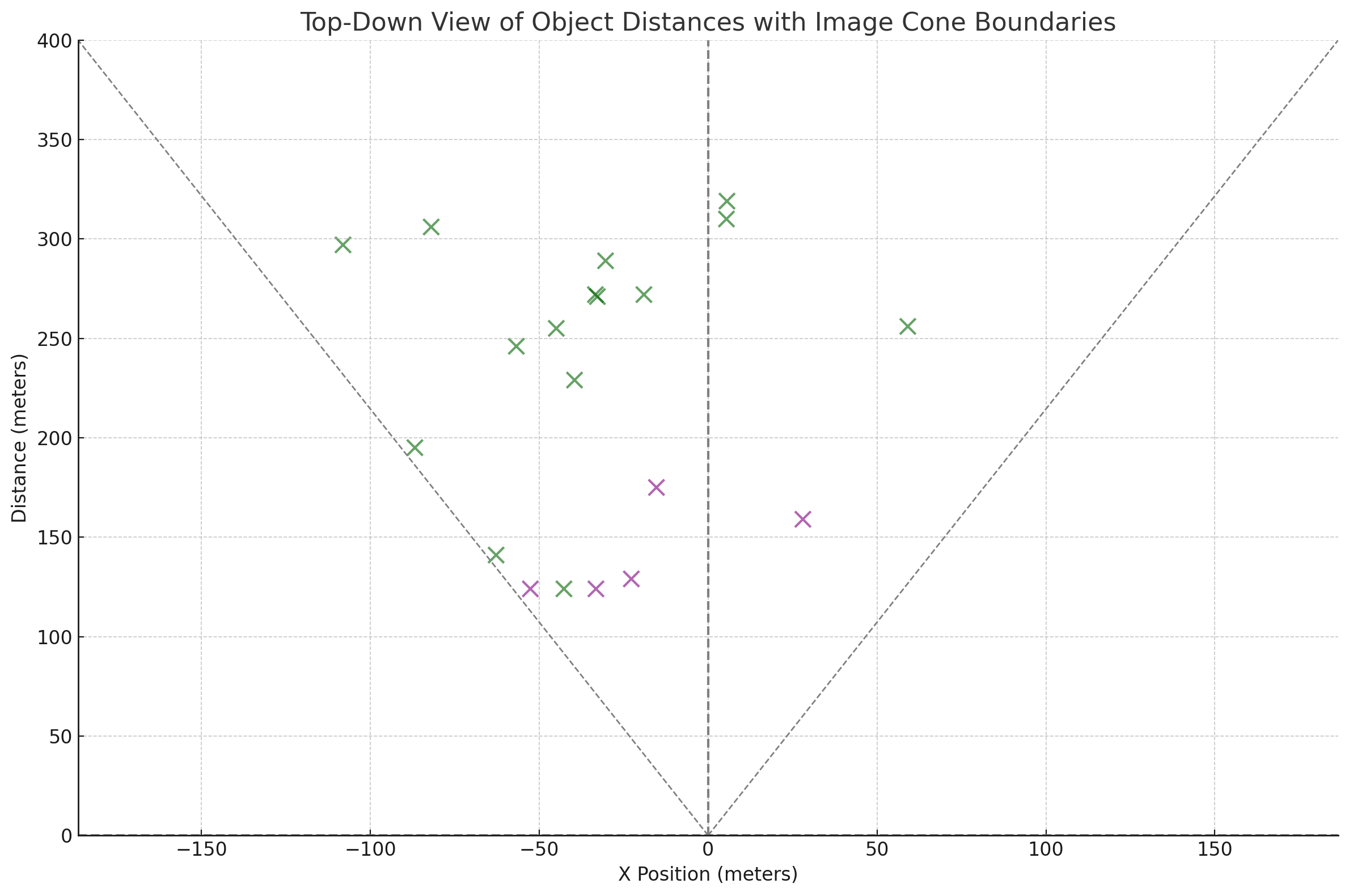}
\caption{Example scene and its bounding box and distance predictions using our method (top). Projected detections onto the 2D plane (bottom).}
\label{fig:distance_predictin_example}
\end{figure}


Unmanned surface vehicles (USVs) and autonomous boats promise efficient autonomous navigation and enhanced safety. However, the development and operational efficiency of USVs are hampered by the limitations of current sensing and perception technologies \cite{liu2016unmanned}.

Traditional sensors for distance estimation, such as LiDAR, radar, sonar, and depth cameras, play a crucial role in navigation and object detection \cite{jorge2019survey}. LiDAR exhibits high costs and sparsity of data. It requires complex processing to interpret sparse point clouds, making it less effective in dynamic, cluttered maritime environments.

Radar systems, while robust in harsh weather conditions, suffer from their own drawbacks. While less expensive than LiDAR, they are still orders of magnitude more expensive than monocular cameras. Their primary technical limitation is the inherent noise in radar signals, which leads to challenges in accurately detecting and classifying small or distant objects \cite{zhuang2016radar}. This limitation is particularly critical in busy maritime channels where the ability to detect small objects can be crucial for collision avoidance.

Sonar sensors are less effective for detecting surface-based obstacles due to their primary design for water depth estimation \cite{bae2015development}. Besides their costliness and limited utility in surface object detection, sonar systems also raise environmental concerns. The acoustic emissions from sonar are known to be disruptive to marine life \cite{abate2010nepa}. These sound waves, often loud, can disturb, disorient, or even harm aquatic animals, especially marine mammals that rely on sound for navigation and communication.

Depth cameras offer a potential alternative, providing rich spatial data about the surroundings. However, they require precise calibration and are constrained by the lack of specialized datasets tailored for marine environments \cite{jorge2019survey}.

Monocular depth estimation leverage neural networks to estimate depth from single images (for each pixel). While promising in theory, this technique is hindered by the quality and variety of available datasets. Existing datasets are predominantly land-based and do not represent the unique challenges of maritime environments, such as water reflections, varying lighting conditions, and diverse objects \cite{yang2024depth}. 

Lastly, trigonometry-based approaches by estimating the orientation of the camera via an onboard IMU (or horizon finding methods), and subsequent ray-casting yield poor results due to the difficulties of obtaining a precise own-orientation in the presence of many different acceleration forces \cite{liu2016unmanned} and as this projection is ill-posed with such acute viewing angles \cite{mou2018wide}. 

On the other hand, we propose a novel approach to approximate distance estimation on USVs using supervised object detection. We circumvent the limitations of traditional sensors by leveraging video-metadata-pairs comprising images with manually annotated bounding boxes and corresponding distance measurements. We obtain the distance measurements by leveraging the onboard CAN bus to get the "NMEA"(National Marine Electronics Association) data stream to collect GPS, heading, and more, and compare to chart data, which we collected from open sources, such as the National Data Buoy Center. Using this as a test dataset, we propose an object detector-based neural network to predict distances of individual objects based on visual cues, akin to human estimations. For that, we also human-label distance by letting annotators gauge distances based on the image context. 

This method has the advantage that any video data can be used to train a system to become better at distance estimation, without the need for any other costly or hard to configure sensors. Moreover, it does not affect the runtime of the underlying object detector, making it a cheap and simple solution.

We demonstrate the practical application of our method by running experiments on embedded hardware and analyzing its performance in combination with multi-object trackers. Our main contributions are as follows:
\begin{itemize}
  \item We propose an approach to approximate distance estimation in USVs using supervised object detection, a cheap and simple method for gauging distances to other objects without adding any runtime requirements.
  \item A comprehensive maritime dataset of manually annotated images with bounding box data and distances is introduced, enabling effective training of a machine learning model for distance estimation. It will be made publicly available.
  \item We analyze the approximation distance estimation approach by comparing it against manually annotated ground truth distances and baseline methods.
\end{itemize}

\section{Related Work}
\label{sec:relatedwork}

Research in the intersection of computer vision and USVs has seen progress in areas, such as horizon estimation \cite{hashmani2020survey,singhal2023marine,kiefer2023fast,fadhil2014runway,liu2020water,wei2016effective,steccanella2020waterline,guo2015fast}, semantic segmentation \cite{qiu2020improved,yang2022image,bovcon2019mastr1325,kristan2015fast,bovcon2021wasr}, panoptic segmentation \cite{vzust2023lars,nirgudkar2023massmind,qiao2022automated}, anomaly/obstacle detection \cite{cane2016saliency,kristan2015fast,prasad2018object}, heading estimation \cite{10256271,captainai_heading_estimation}, and more. Please see the survey or workshop papers on the current progress of maritime computer vision \cite{Kiefer_2023_WACV,Kiefer_2024_WACV,prasad2017video,lyu2022sea,steiniger2022survey,saleh2022computer}.

The field of object detection has matured within the last years from CNN-based two-stage detectors \cite{ren2015faster,carranza2020performance,zou2023object}, one-stage detectors \cite{ouchra2021object,diwan2023object,vijayakumar2024yolo}, up to transformer-based detectors \cite{arkin2023survey,li2023transformer,amjoud2023object}. While there are attempts on making transformer-based detectors embedded-friendly \cite{chen2022mobile,ma2022mocovit,mehta2021mobilevit}, the default choice for embedded detectors are CNN-based one-stage detectors, such as the YOLO series \cite{zhou2024efficient}. In  our experiments, we mostly rely on these as they are the most prevalent in this domain. 
In the maritime domain, object detection has often been connected to obstacle detection for downstream obstacle avoidance \cite{jiang2024obstacle,jin2024wide,Kiefer_2023_WACV}. We refer to \cite{Kiefer_2024_WACV} for a discussion on challenges in maritime object detection. 

Since many vessels don't employ an automatic identification system (AIS) and it also only transponds signals slowly, the need of distance estimation through other means is inevitable \cite{qu2023improving}. There are many radar-based approaches, focusing on reducing the noise or classifying radar blobs \cite{li2019small,del2018artificial,almeida2009radar}. However, radar suffers from the usual challenges of reflectivity and lack of interpretability \cite{robinette2019sensor}. LiDAR-based approaches are investigated as well, but lack the resolution for wider distances \cite{ferreira2022lidar,zha2022research,han2017persistent}. Both sensor types are considerably more expensive than cameras. 

Hence, vision-based solutions are in the main focus of current USV-based computer vision research. Traditional approaches are triangulation-based, e.g.
stereo-vision approaches were investigated in \cite{sinisterra2014stereo,kaplowitz2022characterization,sinisterra2017usv}, where depth information is obtained by comparing the disparity between two horizontally displaced cameras. This method provides accurate distance estimation for objects at varying ranges, but the accuracy diminishes with increasing distance due to reduced disparity. E.g., experiments on KITTI show a poor performance for distances beyond $\sim$100m \cite{pinggera2014know}. Also, the maritime domain has mostly poor features, often only showing moving water and sky. Furthermore, stereo cameras are prone to misconfiguration and very sensitive to calibration. 

Traditional monocular  approaches rely on geometric triangulation based on knowing the ego-pose and camera intrinsics \cite{mousazadeh2018developing,10256271}. However, these approaches are ill-posed with small variations in self-orientation estimation leading to large variations in pixel space \cite{ali2020real}. \cite{mansour2019relative,liu2022vision} show that parallax-based approaches can outperform stereo vision for greater distances, but it remains inaccurate for larger distances.

Lately, fully monocular depth estimation approaches by means of end to end neural networks have been investigated as a means of gauging distances \cite{ming2021deep,bhoi2019monocular,yang2024depth}. While promising, these approaches rely on large amounts of accurate depth maps, either from stereo cameras, lidar, or synthetic data. Furthermore, they tend to work in indoor scenes with limited distances only \cite{roussel2019monocular}.

Contrary to these approaches, we focus on approximate object-based supervised distance estimation \cite{davydov2022supervised}. The maritime domain has unique challenges and our approach has the benefit to work with larger distances, can be trained fully end to end without any considerable overhead and is simple to integrate into existing object detection architectures. Please see Section \ref{sec:results} for an analysis and comparison of different methods.

Lastly, to facilitate this research direction, we collected and annotated a large dataset with bounding box labels accompanied with distances. Current maritime CV datasets focus on object detection, multi-object tracking, semantic segmentation  \cite{su2023survey} or focus on other sensor modalities \cite{de2010dataware}. However, CV-based distance estimation is a relatively young domain without any publicly available datasets.

\begin{figure*}
\centering
\includegraphics[width=1\textwidth]{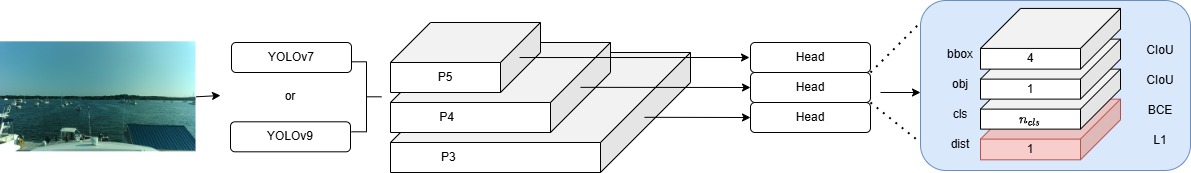}
\caption{Architecture of our proposed approach at the example of YOLOv7 and YOLOv9. We leave the base architecture of the networks the same except for adding a distance loss branch to the heads (in red).}
\label{fig:yolodistances}
\end{figure*}

\section{Methodology}
\label{sec:methodology}

\subsection{Model Architecture}
We employ the YOLOv7 and YOLOv9 series of object detectors for our task. These detectors are known for being on the Pareto front of accuracy and real-time performance \cite{wang2024yolov9}. To adapt these models for distance estimation, we modify the architecture to predict an additional output for distance as part of each anchor as shown in Figure \ref{fig:yolodistances}.

The adjusted model architecture integrates distance prediction by including an extra output neuron for each anchor. This additional output allows the model to predict the distance to the detected object directly. This formulation allows us 

However, these networks predict numbers that are "well-behaved", in that they are mostly centered around zero and are of low magnitude \cite{wang2024yolov9}. Directly predicting the metric distances would result in unstable training or very poor performance. Hence, we experimented with various normalization strategies for the distance prediction branch, including linear scaling, logarithmic scaling, and hybrid approaches to handle different ranges and distributions of distances.

\subsection{Distance Normalization Strategies}

We explored four main normalization strategies for the distance prediction:

\subsubsection{Linear Normalization}

The linear normalization scales the actual distance \(d\) to a normalized value \(y\) in the range \([0,1]\):

\begin{equation}
y = \frac{d}{d_{\text{max}}}
\end{equation}

During inference, we recover the predicted distance \(\hat{d}\) from the network's output \(y\):

\begin{equation}
\hat{d} = y \times d_{\text{max}}
\end{equation}

\subsubsection{Logarithmic Normalization}

For logarithmic normalization, we apply a logarithmic transformation and scale to \([0,1]\):

\begin{equation}
y = \frac{\log(d + \epsilon)}{\log(d_{\text{max}} + \epsilon)}
\end{equation}

Here, \(\epsilon\) is a small positive constant to avoid taking the logarithm of zero (e.g., \(\epsilon = 1\)).

During inference:

\begin{equation}
\hat{d} = \exp\left( y \times \log(d_{\text{max}} + \epsilon) \right) - \epsilon
\end{equation}

\subsubsection{Linear Negative Normalization}

To map distances to the range \([-1,1]\):

\begin{equation}
y = 2 \left( \frac{d}{d_{\text{max}}} \right) - 1
\end{equation}

During inference:

\begin{equation}
\hat{d} = \left( \frac{y + 1}{2} \right) \times d_{\text{max}}
\end{equation}

\subsubsection{Logarithmic Negative Normalization}

Combining logarithmic scaling with mapping to \([-1,1]\):

\begin{equation}
y = 2 \left( \frac{\log(d + \epsilon)}{\log(d_{\text{max}} + \epsilon)} \right) - 1
\end{equation}

During inference:

\begin{equation}
\hat{d} = \exp\left( \left( \frac{y + 1}{2} \right) \times \log(d_{\text{max}} + \epsilon) \right) - \epsilon
\end{equation}

\subsection{Training and Inference}

By applying these normalization strategies, we ensure that the network's distance predictions \(y\) are within ranges suitable for neural network outputs. During training, the loss is computed between the normalized predicted distances and the normalized ground truth distances. During inference, the inverse transformations are applied to map the network's outputs back to actual distance values \(\hat{d}\).

\subsection{Model Training}
The model is trained using a composite loss function that balances several components:
\begin{itemize}
  \item \textbf{Objectness Loss}: Penalizes incorrect predictions of object presence.
  \item \textbf{Classification Loss}: Ensures accurate classification of detected objects.
  \item \textbf{Localization Loss}: Measures the accuracy of bounding box predictions.
  \item \textbf{Distance Loss}: Evaluates the accuracy of the distance predictions.
\end{itemize}

We leave the loss functions for the first three losses unchanged, and use $L_1$ as the distance loss.

\subsection{Dataset Collection and Annotation}

\begin{figure}
\centering
\includegraphics[width=0.5\textwidth]{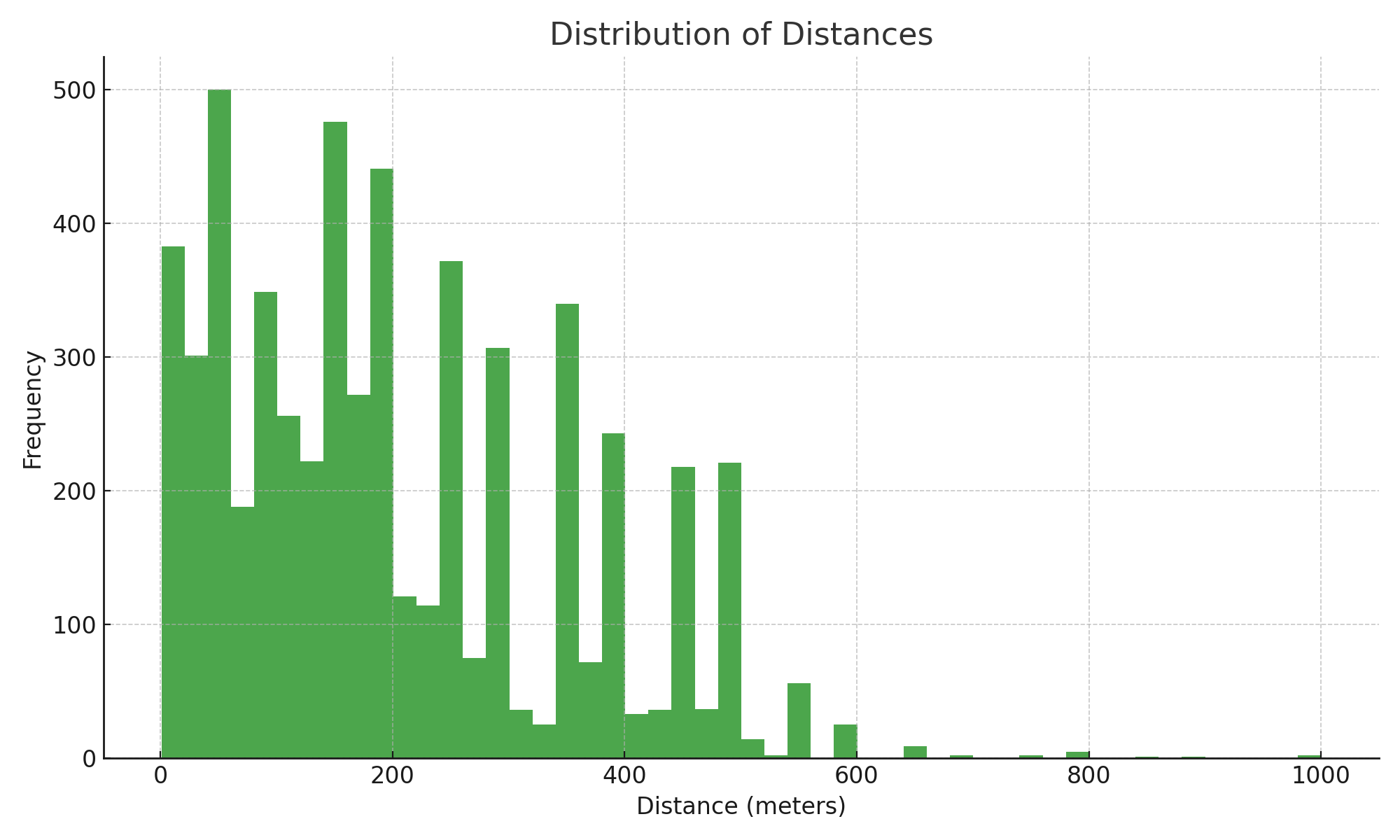}
\caption{Distribution of distances in our dataset.}
\label{fig:distance_distribution}
\end{figure}

\begin{figure}
\centering
\includegraphics[width=0.49\textwidth]{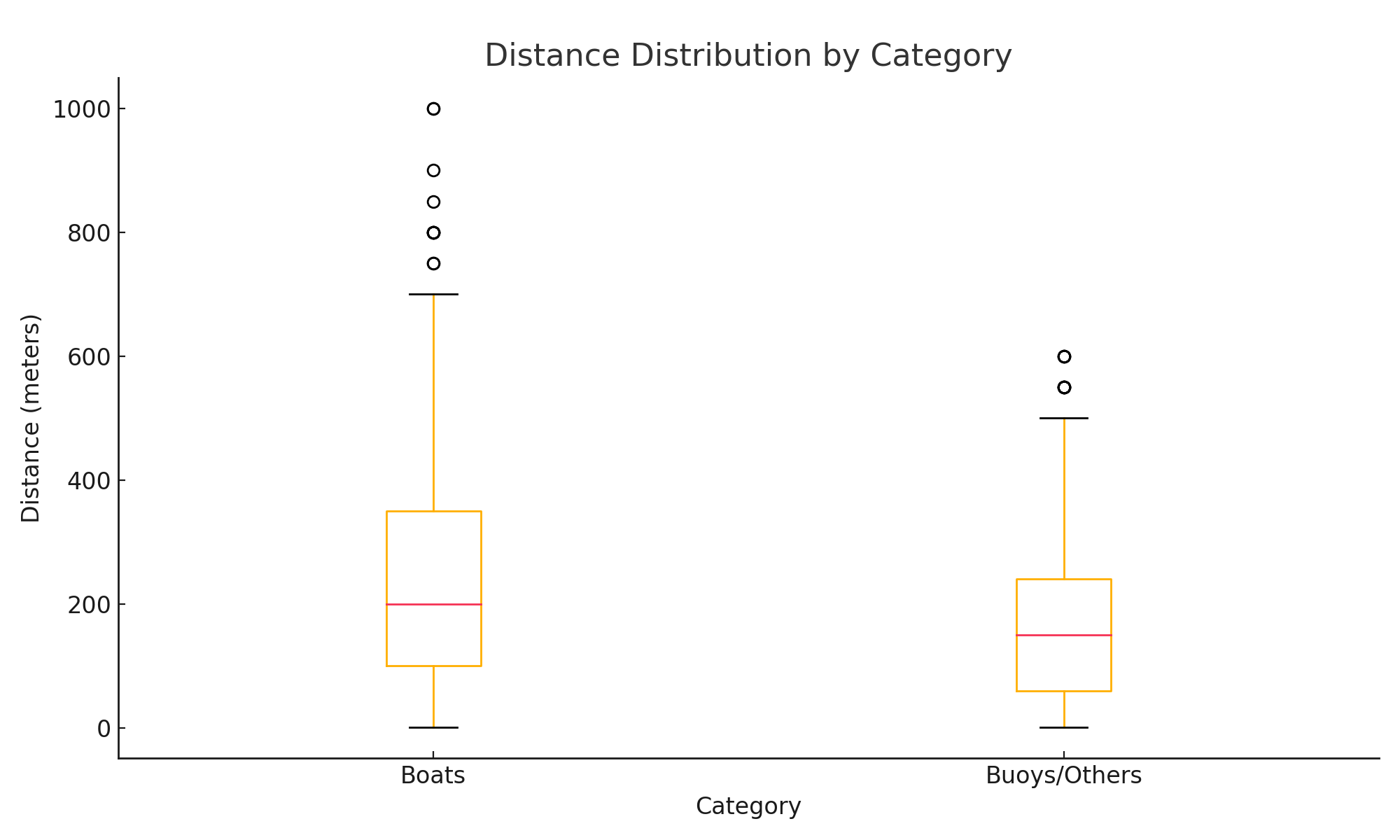}
\caption{Distance Distribution by Category}
\label{fig:distance_boxplot}
\end{figure}

Our evaluation dataset includes images captured from USVs, manually annotated with bounding boxes and distances. We captured the images using rented boats on the east coast of the US. For ground truth evaluation (only for that - for our proposed method, we use manually labelled data), we determined distances to static objects such as buoys and docked boats using the USV's GPS position and heading and verified manually with known chart positions. For the chart data, we manually downloaded and fused buoy data for the US from NOAH. 

Our evaluation dataset consists of 1,000 images captured from USVs in various maritime environments, including open waters, harbors, and coastal regions. Each image is annotated with bounding boxes around objects such as boats, buoys, and other obstacles, along with their respective distances from the USV. Distances are measured using GPS positioning for static objects. This comprehensive dataset ensures robust training and evaluation of our model.

\begin{figure*}
\centering
\includegraphics[width=1\textwidth]{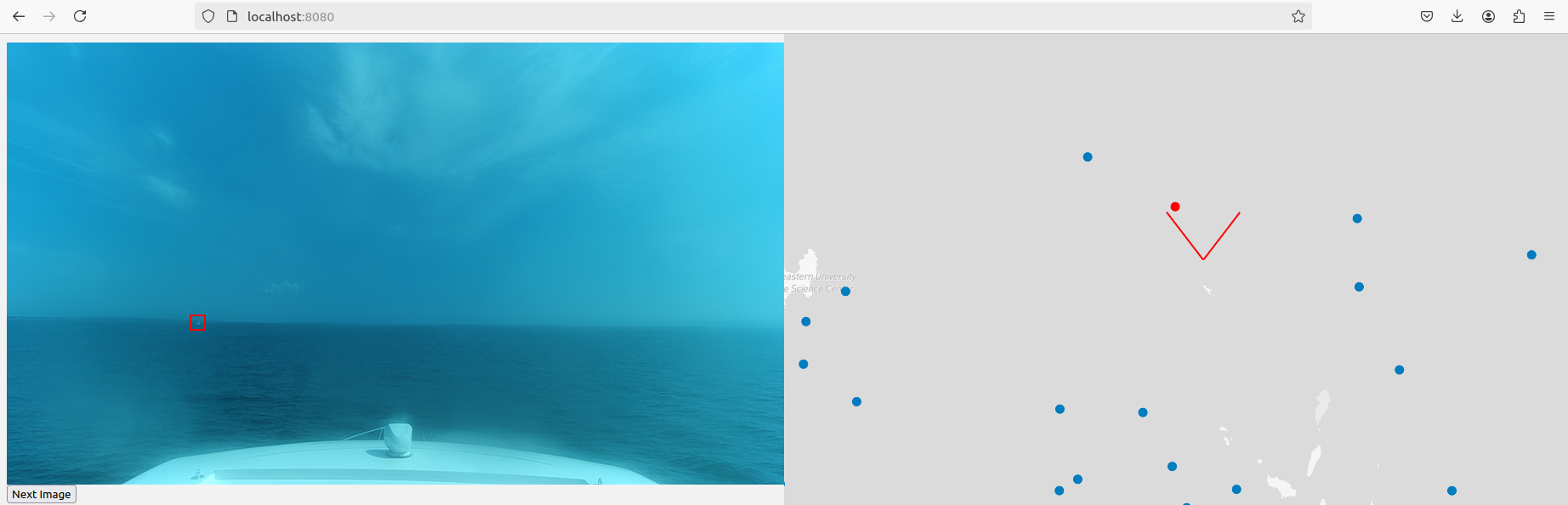}
\caption{Labeling tool used for getting bounding boxes and associating chart data with vision data.}
\label{fig:labelingtool}
\end{figure*}

We created our own labeling tool, which combines all these components: bounding box labeling, accompanying chart data integration, and association labelling. See Figure \ref{fig:labelingtool} for a screenshot from its UI. We're making it publicly available. Also see Figure \ref{fig:distance_distribution} for distance histogram. 

\subsection{Human Labeling For More Data}

To avoid the need for metadata alongside the image data, we also explored the approach of having human annotators gauge distances to objects. For this, we asked a professional labeling service company to label bounding boxes and alongside of it distances. With this approach, we obtained another 5,000 images with pseudo-distances. We'll explore in the experiment section how useful this data is.

\section{Evaluation Metrics}
\label{sec:evaluation_metrics}

To evaluate the performance of our object detection and distance estimation model, we use two primary metrics: mean Average Precision (mAP) and a novel weighted distance error metric. We calculate mean Average Precision (mAP) using an Intersection over Union (IoU) threshold of 0.5 to determine correct detections.

\subsection{Weighted Distance Error Metric}
The weighted distance error metric evaluates the accuracy of the distance predictions while considering the confidence of each detection. For each detected object \( i \), let \( d_i \) be the ground truth distance, \( \hat{d}_i \) be the predicted distance, and \( c_i \) be the confidence score of the detection. The distance error \( e_i \) is given by $e_i = |d_i - \hat{d}_i|$.

To weight the distance error by the confidence, we define the weighted distance error \( E \) as:

\[
E = \frac{\sum_{i=1}^M c_i \cdot e_i}{\sum_{i=1}^M c_i}
\]

where \( M \) is the total number of detections. This metric ensures that higher confidence detections have a greater impact on the overall error, reflecting their presumed accuracy.

Similar to the mAP calculation, we only consider distance errors for detections whose IoU with the ground truth exceeds a predefined threshold. This ensures that only sufficiently accurate object detections contribute to the weighted distance error metric.

\section{Experiments}
\label{sec:results}

In this section, we evaluate the performance of our proposed approach for approximate distance estimation on unmanned surface vehicles (USVs). We conduct experiments to assess both object detection accuracy and distance estimation precision. The evaluation includes comparisons between different model architectures, analysis of distance estimation errors across distance intervals and object types, and benchmarking against other distance estimation methods. All experiments are performed using our maritime dataset described in Section~\ref{sec:methodology}.

\subsection{Object Detection Performance}

We first evaluate the object detection capabilities of our modified YOLO models. Table~\ref{tab:od_performance} presents the precision, recall, mean Average Precision at IoU threshold 0.5 (mAP@0.5), mean Average Precision across IoU thresholds from 0.5 to 0.95 (mAP@0.5:0.95), and inference speed measured in frames per second (FPS). The models are tested on an NVIDIA Orin AGX using the DeepStream toolkit, with speed measured as wall-clock time from camera input to display output.

As shown in Table~\ref{tab:od_performance}, the larger models (YOLOv7 and YOLOv9-M) achieve higher detection accuracy, with YOLOv9-M attaining the highest mAP@0.5 of 81\%. The tiny models offer faster inference speeds, with YOLOv9-Tiny reaching up to 72.8 FPS, making them suitable for real-time applications where computational resources are limited.

Please see Figure \ref{fig:distance_predictin_example} for a qualitative result of YOLOv7 with integrated distance estimation.

\begin{table} 
\centering
\caption{Object detection performance and speed as measured on an NVIDIA Orin AGX, using the DeepStream toolkit. Speed is in wall-clock time from camera to display.}
\label{tab:od_performance}
\begin{tabular}{lccccc}
\hline
\textbf{Metric} & \textbf{Pr} & \textbf{Re} & \textbf{mAP@0.5} & \textbf{mAP@0.5:0.95} & \textbf{FPS} \\
\hline
YOLOv7-T & 0.82 & 0.68 & 0.73 & 0.45 & 56.5 \\
YOLOv7      & 0.85 & 0.72 & 0.78 & 0.50 & 45.2 \\
YOLOv9-T    & 0.83 & 0.70 & 0.76 & 0.48 & 72.8 \\
YOLOv9-M    & 0.87 & 0.74 & 0.81 & 0.53 & 48.8 \\
\hline
\end{tabular}
\end{table}

\subsection{Distance Estimation Accuracy}

To assess the distance estimation performance, we analyze the weighted distance error across different distance intervals and object types. Before selecting the normalization strategy for the distance prediction, we conducted preliminary tests with the different normalization methods described in Section~\ref{sec:methodology}, including linear normalization, logarithmic normalization, linear negative normalization, and logarithmic negative normalization. We found that only the basic linear normalization yielded satisfactory results. The other methods either did not converge during training or resulted in unstable predictions. Therefore, we proceeded with the basic linear normalization in our experiments.

Table~\ref{tab:mean_dist_errors_narrow} presents the weighted distance errors (in meters) for boats and buoys over distance ranges from 0 to 500 meters (clipped), using different model architectures.

\begin{table}[!ht]
\centering
\footnotesize
\setlength{\tabcolsep}{3pt} 
\caption{Weighted distance error by distance interval (in meters) and object type (\textbf{b}oa\textbf{t}s and \textbf{b}uo\textbf{y}s) for different models.}
\label{tab:mean_dist_errors_narrow}
\begin{tabular}{lcccccccc}
\hline
\multirow{2}{*}{\shortstack{\textbf{Dist.}\\($\times 100$)}} 
  & \multicolumn{2}{c}{\textbf{YOLOv7-t}}
  & \multicolumn{2}{c}{\textbf{YOLOv7}}
  & \multicolumn{2}{c}{\textbf{YOLOv9-t}}
  & \multicolumn{2}{c}{\textbf{YOLOv9-m}} \\
& \textbf{Bt} & \textbf{By} 
  & \textbf{Bt} & \textbf{By}
  & \textbf{Bt} & \textbf{By}
  & \textbf{Bt} & \textbf{By} \\
\hline
$0$-$1$  & 19.1 & 16.0 & 18.2 & 15.1 & 20.4 & 17.0 & 18.7 & 15.8 \\
$1$-$2$  & 39.6 & 33.3 & 37.8 & 32.0 & 41.1 & 34.9 & 38.5 & 33.1 \\
$2$-$3$  & 47.7 & 55.5 & 45.9 & 53.6 & 50.2 & 57.0 & 48.1 & 54.8 \\
$3$-$4$  & 63.4 & 91.1 & 61.2 & 89.1 & 65.7 & 92.6 & 62.8 & 90.4 \\
$4$-$5$  & 76.7 & 131.2 & 74.6 & 129.0 & 78.4 & 133.6 & 76.0 & 130.9 \\
\hline
\textbf{Avg.} & 49.3 & 65.4 & 47.5 & 63.8 & 51.2 & 67.0 & 48.8 & 65.0 \\
\hline
\end{tabular}
\end{table}

The results indicate that the distance estimation error increases with the distance to the object, which is expected due to reduced visual cues at greater distances. Our models consistently perform better on boats than buoys, likely because boats are larger and have more distinct features. Among the models, YOLOv7 and YOLOv9-M achieve lower average distance errors, demonstrating the effectiveness of our approach in accurately estimating distances.

\subsection{Comparison with Other Methods}

We compare our proposed method with traditional triangulation-based distance estimation (based on orientation via pitch and roll) and state-of-the-art monocular depth estimation models (dense distance values are projected into bounding box by taking the median distance within a box). Table~\ref{tab:comparison_methods_concrete} summarizes the mean distance error (MDE) in meters, the percentage of outliers, and the inference speed in FPS for each method.

\begin{table} 
\centering
\caption{Comparison of Distance Estimation Methods Based on Ground Truth Data for USVs.}
\label{tab:comparison_methods_concrete}
\begin{tabular}{lcccc}
\hline
\textbf{Method} & \textbf{MDE (m)} & \textbf{Outl. (\%)} & \textbf{FPS} \\
\hline
YOLOv7 + Triangulation & 35.2 & 40.0  & 45.2 \\
Depth Anything V2 (Small) & 43.5 & 50.0  & 10 \\
Depth Anything V2 (Base) & 27.1 & 37.5 &  2 \\
Depth Anything V2 (Large) & 19.2 & 25.0 & 0.5 \\
\textbf{Proposed YOLOv7} & 14.9 & 12.0 &  45.2 \\
\hline
\end{tabular}
\end{table}

Our proposed method achieves the lowest mean distance error and the smallest percentage of outliers while maintaining real-time performance. Traditional triangulation methods and monocular depth estimation models either have higher errors or are computationally intensive, making them less suitable for real-time USV applications. Depth Anything v2 struggles with distances beyond approx. 60m as it is not trained to do so.

\subsection{Loss Balancing and Ablation Studies}
We conducted ablation studies to understand the impact of the distance loss gain on the model's performance for both object detection (OD) and distance estimation. As shown in Table \ref{tab:loss_weights_distance}, the distance loss gain significantly affects the balance between OD accuracy and distance estimation precision.

\begin{table}
\centering
\caption{Ablation Studies on Distance Loss Gain Settings (Other loss components are fixed: Box Loss = 0.05, Classification Loss = 0.3, Objectness Loss = 0.7)}
\label{tab:loss_weights_distance}
\begin{tabular}{lccc}
\hline
\textbf{Distance Loss Gain (distance)} & \textbf{0.001} & \textbf{0.01} & \textbf{0.1} \\
\hline
mAP@0.5 (OD) & 89.5\% & 86.4\% & 72.1\% \\
MAE (m) (Distance) & 19.7 & 14.2 & 9.8 \\
\hline
\end{tabular}
\end{table}

The results indicate that increasing the distance loss gain improves distance estimation accuracy (lower MAE) but significantly reduces OD performance (lower mAP). A lower distance loss gain (e.g., 0.001) maintains higher OD accuracy but at the cost of less precise distance predictions. Balancing these loss components is crucial for optimizing both OD and distance estimation tasks.

\subsection{Smoothing the Distance Estimates}
\label{sec:video_distance_estimation}

To handle distance estimation in video sequences, where objects move and change positions over time, we use the Simple Online and Realtime Tracking (SORT) algorithm \cite{Bewley2016_sort}. SORT tracks objects across frames by combining object detection results with a Kalman filter and the Hungarian algorithm for data association, ensuring that each detected object maintains a consistent identity across frames.

Alongside tracking, we keep a running average of the distance estimates for each tracked object. This simple averaging technique helps to smooth the distance predictions over time, filtering out outliers caused by sudden detection errors or rapid changes in object appearance. The approach is computationally efficient and provides stable distance estimates, which are crucial for downstream tasks like collision avoidance in dynamic maritime environments.

By combining SORT with a running average for distance smoothing, we achieve robust and real-time performance on Unmanned Surface Vehicles (USVs), ensuring that distance predictions are both accurate and temporally consistent. Table \ref{tab:tracking_comparison} shows how tracking and smoothing reduces the Mean Distance Error (MDE) and the percentage of outliers.

\begin{table} 
\centering
\caption{Impact of Tracking and Smoothing on Distance Estimation Performance for YOLOv9-M.}
\label{tab:tracking_comparison}
\begin{tabular}{lccc}
\hline
\textbf{Method} & \textbf{MDE (m)} & \textbf{Outl. (\%)} & \textbf{FPS} \\
\hline
Without Tracking & 18.4 & 22.5 & 48 \\
With Tracking (SORT + Avg.) & 14.9 & 12.0 & 45 \\
\hline
\end{tabular}
\end{table}

\subsection{Evaluation of Human-Labeled Distances}
\label{sec:human_vs_chart_evaluation}

To assess the accuracy of human-labeled distances compared to chart-derived ground truth data, we examine human estimations of distances to boats and buoys in maritime settings. The errors in human distance estimation generally increase with the distance to the object.

We categorized the gt-labeled data, which has been labeled manually as well, into three ranges: Close (0-100 m), Medium (100-300 m), and Far (300+ m). For each range, we computed the Mean Absolute Error (MAE) and Mean Absolute Percentage Error (MAPE) to quantify deviations from chart data. The results are summarized in Table \ref{tab:human_vs_chart_data}.

\begin{table}
\centering
\caption{Distance Estimation Errors for Human-Labeled Ground Truth Compared to Chart-Based Ground Truth by Object Type and Distance Range.}
\label{tab:human_vs_chart_data}
\begin{tabular}{lcc|cc}
\hline
\textbf{Dist} & \multicolumn{2}{c|}{\textbf{Boats}} & \multicolumn{2}{c}{\textbf{Buoys}} \\
\textbf{Range} & \textbf{MAPE \%} & \textbf{MAE (m)} & \textbf{MAPE \%} & \textbf{MAE (m)} \\
\hline
0-100  & 45.2 & 17.8 & 52.7 & 20.4 \\
100-300 & 64.9 & 48.3 & 70.3 & 56.1 \\
300+ & 84.1 & 98.2 & 89.2 & 115.7 \\
\hline
\end{tabular}
\end{table}

Table \ref{tab:human_vs_chart_data} shows human-labeled distance estimates are more accurate at closer ranges, with lower MAE and MAPE. However, accuracy decreases for medium and far ranges. For example, at distances over 300 m, the MAE for boats and buoys exceeds 98 m and 115 m, respectively, with MAPE approaching 90\%. Notably, for various use-cases it is not important to get a precise distance estimate if an object is far away.

\section{Discussion and Limitations}
\label{sec:discussion}

Our method provides a cost-effective alternative to traditional sensors for USVs, enhancing safety and operational efficiency. However, there are limitations to consider.

A key limitation is the sensitivity of our distance estimation to changes in the camera's field of view (FoV) and other camera parameters like focal length and mounting position. The model learns to associate object sizes with distances based on the training FoV. If deployed with a different FoV or zoom level, the model could produce inaccurate estimates, as it cannot inherently differentiate between varying FoVs. For instance, a narrower FoV might make distant objects appear closer, leading to erroneous distance predictions. Future work could address this by incorporating FoV information or calibration steps to ensure consistent distance estimation across different camera setups.

\section{Conclusion}

We have introduced a novel approach for approximate distance estimation on USVs using supervised object detection, providing a cost-efficient alternative to traditional sensors. Our method achieves solid performance in both object detection and distance estimation while maintaining real-time speeds suitable for USV applications. Future work may focus on handling varying camera parameters and environmental conditions to enhance robustness.

\newpage
\bibliographystyle{IEEEtran}
\bibliography{IEEEabrv,IEEEexample}

\end{document}